%% file: paper.tex
\documentclass[10pt,twocolumn,letterpaper]{article}

\usepackage[pagenumbers]{cvpr} 

\usepackage{graphicx}
\usepackage{amsmath}
\usepackage{amssymb}
\usepackage{booktabs}
\usepackage{upgreek}
\usepackage{tikz}
\usepackage{comment}
\usepackage{color}
\usepackage{bigstrut}
\usepackage{algorithm}
\usepackage{algorithmic}
\usepackage{multirow}
\usepackage{bbm}
\usepackage{booktabs}
\usepackage{enumerate}
\usepackage[misc]{ifsym}
\usepackage{colortbl}
\usepackage{xcolor}
\usepackage{array}
\usepackage{wrapfig}

\usepackage[accsupp]{axessibility}  

\usepackage[pagebackref,breaklinks,colorlinks]{hyperref}

\usepackage[capitalize]{cleveref}
\crefname{section}{Sec.}{Secs.}
\Crefname{section}{Section}{Sections}
\Crefname{table}{Table}{Tables}
\crefname{table}{Tab.}{Tabs.}

\def\cvprPaperID{****} 
\def\confName{CVPR}
\def\confYear{2023}

\begin{document}

\title{RepMode: Learning to Re-parameterize Diverse Experts \\ for Subcellular Structure Prediction}

\author{
Donghao Zhou \textsuperscript{1, 2, 3} \and
Chunbin Gu \textsuperscript{3} \and
Junde Xu \textsuperscript{1, 2, 3} \and
Furui Liu \textsuperscript{4} \and
Qiong Wang \textsuperscript{1} \and 
Guangyong Chen \textsuperscript{4}\thanks{Corresponding Author: gychen@zhejianglab.com} \and
Pheng-Ann Heng \textsuperscript{1, 3, 4} \vspace{1.5mm} \and
\textsuperscript{1} Guangdong Provincial Key Laboratory of Computer Vision and Virtual Reality Technology, \\ Shenzhen Institute of Advanced Technology, Chinese Academy of Sciences \and
\textsuperscript{2} University of Chinese Academy of Sciences \and
\textsuperscript{3} The Chinese University of Hong Kong \and
\textsuperscript{4} Zhejiang Lab 
\vspace{1.5mm} \\ \normalsize{\url{https://correr-zhou.github.io/RepMode}}
}

\maketitle



\begin{abstract}

    In biological research, fluorescence staining is a key technique to reveal the locations and morphology of subcellular structures.
    However, it is slow, expensive, and harmful to cells.
    In this paper, we model it as a deep learning task termed subcellular structure prediction (SSP), aiming to predict the 3D fluorescent images of multiple subcellular structures from a 3D transmitted-light image.
    Unfortunately, due to the limitations of current biotechnology, each image is partially labeled in SSP.
    Besides, naturally, subcellular structures vary considerably in size, which causes the multi-scale issue of SSP.
    To overcome these challenges, we propose Re-parameterizing Mixture-of-Diverse-Experts (RepMode), a network that dynamically organizes its parameters with task-aware priors to handle specified single-label prediction tasks.
    In RepMode, the Mixture-of-Diverse-Experts (MoDE) block is designed to learn the generalized parameters for all tasks, and gating re-parameterization (GatRep) is performed to generate the specialized parameters for each task, by which RepMode can maintain a compact practical topology exactly like a plain network, and meanwhile achieves a powerful theoretical topology. 
    Comprehensive experiments show that RepMode can achieve state-of-the-art overall performance in SSP.
    
\end{abstract}

\section{Introduction}

\input{figures/1-SSP}

Recent years have witnessed great progress in biological research at the subcellular level \cite{thul2017subcellular, christopher2021subcellular, gut2018multiplexed, carlton2020membrane, wolff2020molecular, guo2018visualizing, bock2020mitochondria}, which plays a pivotal role in deeply studying cell functions and behaviors.
To address the difficulty of observing subcellular structures, fluorescence staining was invented and has become a mainstay technology for revealing the locations and morphology of subcellular structures \cite{herman2020fluorescence}.
Specifically, biologists use the antibodies coupled to different fluorescent dyes to ``stain'' cells, after which the subcellular structures of interest can be visualized by capturing distinct fluorescent signals \cite{xu2022deep}.
Unfortunately, fluorescence staining is expensive and time-consuming due to the need for advanced instrumentation and material preparation \cite{im2019introduction}. Besides, phototoxicity during fluorescent imaging is detrimental to living cells \cite{icha2017phototoxicity}.
In this paper, we model fluorescence staining as a deep learning task, termed \textit{subcellular structure prediction} (\textit{SSP}), which aims to directly predict the 3D fluorescent images of multiple subcellular structures from a 3D transmitted-light image (see \cref{fig:SSP}(a)). 
The adoption of SSP can significantly reduce the expenditure on subcellular research and free biologists from this demanding workflow.

Such an under-explored and challenging bioimage problem deserves the attention of the computer vision community due to its high potential in biology.
Specifically, SSP is a dense regression task where the fluorescent intensities of multiple subcellular structures need to be predicted for each transmitted-light voxel.
However, due to the limitations of current biotechnology, each image can only obtain \textit{partial labels}. 
For instance, some images may only have the annotations of nucleoli, and others may only have the annotations of microtubules (see \cref{fig:SSP}(b)). 
Moreover, different subcellular structures would be presented at \textit{multiple scales} under the microscope, which also needs to be taken into account.
For example, the mitochondrion is a small structure inside a cell, while obviously the cell membrane is a larger one since it surrounds a cell (see \cref{fig:SSP}(c)). 

Generally, there are two mainstream solutions:
1) \textit{Multi-Net} \cite{ounkomol2018label, jo2021label, cheng2021single, kandel2020phase}: divide SSP into several individual prediction tasks and employs multiple networks;
2) \textit{Multi-Head} \cite{christiansen2018silico, cross2022label, manifold2021versatile}: design a partially-shared network composed of a shared feature extractor and multiple task-specific heads (see \cref{fig:overview}(a)).
However, these traditional approaches organize network parameters in an \textit{inefficient} and \textit{inflexible} manner, which leads to two major issues.
First, they fail to make full use of partially labeled data in SSP, resulting in \textit{label-inefficiency}. 
In Multi-Net, only the images containing corresponding labels would be selected as the training set for each network and thus the other images are wasted, leading to an unsatisfactory generalization ability. 
As for Multi-Head, although all images are adopted for training, only partial heads are updated when a partially labeled image is input and the other heads do not get involved in training.
Second, to deal with the multi-scale nature of SSP, they require exhausting pre-design of the network architecture, and the resultant one may not be suitable for all subcellular structures, which leads to \textit{scale-inflexibility}.

In response to the above issues, herein we propose \textit{Re-parameterizing Mixture-of-Diverse-Experts} (\textit{RepMode}), an all-shared network that can dynamically organize its parameters with task-aware priors to perform specified single-label prediction tasks of SSP (see \cref{fig:overview}(b)).
Specifically, RepMode is mainly constructed of the proposed \textit{Mixture-of-Diverse-Experts} (\textit{MoDE}) \textit{blocks}.
The MoDE block contains the expert pairs of various receptive fields,
where these \textit{task-agnostic} experts with diverse configurations are designed to \textit{learn the generalized parameters for all tasks}.
Moreover, \textit{gating re-parameterization} (\textit{GatRep}) is proposed to conduct the \textit{task-specific} combinations of experts to achieve efficient expert utilization, which aims to \textit{generate the specialized parameters for each task}.
With such a parameter organizing manner (see \cref{fig:overview}(c)), RepMode can maintain a practical topology exactly like a plain network, and meanwhile achieves a theoretical topology with a better representational capacity. 
Compared to the above solutions, RepMode can fully learn from all training data, since the experts are shared with all tasks and thus participate in the training of each partially labeled image.
Besides, RepMode can adaptively learn the preference of each task for the experts with different receptive fields, thus no manual intervention is required to handle the multi-scale issue. 
Moreover, by fine-tuning few newly-introduced parameters, RepMode can be easily extended to an unseen task without any degradation of the performance on the previous tasks.
Our main contributions are summarized as follows:
\begin{itemize}
    \setlength{\itemsep}{0pt}
    \setlength{\parskip}{3pt}
    \item We propose a stronger baseline for SSP, named RepMode, which can switch different ``modes'' to predict multiple subcellular structures and also shows its potential in task-incremental learning.
    \item The MoDE block is designed to enrich the generalized parameters and GatRep is adopted to yield the specialized parameters, by which RepMode achieves dynamic parameter organizing in a task-specific manner.
    \item Comprehensive experiments show that RepMode can achieve state-of-the-art (SOTA) performance in SSP. Moreover, detailed ablation studies and further analysis verify the effectiveness of RepMode.
\end{itemize}


\input{figures/2-overview}

\section{Related Works}
\label{sec:related_work}

\textbf{Partially labeled dense prediction.}
In addition to SSP, many other dense prediction tasks could also face the challenge of partial labeling.
In general, the previous methods can be divided into two groups.
The first one seeks for an effective training scheme by adopting knowledge distillation \cite{feng2021ms, zhang2022unsupervised}, learning cross-task consistency \cite{li2022learning}, designing jointly-optimized losses \cite{shi2021marginal}, \etc.
The second one aims to improve the network architecture with a dynamic segmentation head \cite{zhang2021dodnet}, task-guided attention modules \cite{wu2022tgnet}, conditional tensor incorporation \cite{dmitriev2019learning}, \etc.
However, these methods are primarily developed for large-scale datasets.
Compared to these well-explored tasks, SSP only has relatively small datasets due to the laborious procedure of fluorescence staining. Thus, the training data of SSP should be utilized in a more efficient way.
In light of that, we adopt a task-conditioning strategy in RepMode, where all parameters are shared and thus can be directly updated using the supervision signal of each label.
Unlike other task-conditional networks \cite{wu2022tgnet, zhang2021dodnet, dmitriev2019learning, sun2021task}, our RepMode is more flexible and capable of maintaining a compact topology.

\textbf{Multi-scale feature learning.}
Multi-scale is a fundamental problem of computer vision, caused by the variety in the size of the objects of interest. 
The common solutions are adopting multi-resolution input \cite{zhang2016joint, fang2020multi, fu2017look, yang2020mutualnet}, designing parallel branches \cite{szegedy2015going, chen2017rethinking, li2019scale, zhao2017pyramid, he2015spatial, li2019selective}, fusing cross-layer features \cite{long2015fully, redmon2018yolov3, liu2018path, sun2019deep}, performing hierarchical predictions \cite{lin2017feature, lin2017focal, liu2016ssd}, \etc.
These methods often adopt a pre-defined architecture for all objects to extract multi-scale features in a unified fashion.
In contrast, RepMode learns the dynamic combinations of the experts with different receptive fields for each subcellular structure, and thus is capable of learning multi-scale features in a task-specific manner.

\textbf{Mixture-of-Experts.} 
Mixture-of-Experts (MoE) typically consists of a gating module and multiple independent learners (\ie experts) \cite{yuksel2012twenty}. 
For an input sample, MoE would adaptively assemble the corresponding output of all experts \cite{ma2018modeling, tang2020progressive, dai2021generalizable, pavlitskaya2020using} or only route it to a few specific experts \cite{shazeer2017outrageously, hazimeh2021dselect, jacobs1991adaptive, riquelme2021scaling}, which depends on the gating strategy. 
Benefiting from its divide-and-conquer principle, MoE is widely adopted in computer vision \cite{dai2021generalizable, riquelme2021scaling, wu2022residual, gross2017hard, pavlitskaya2020using}, natural language processing \cite{shazeer2017outrageously, gururangan2021demix, clark2022unified}, and recommendation systems \cite{ma2018modeling, tang2020progressive, qin2020multitask, hazimeh2021dselect}.
Our RepMode is established based on the idea of MoE, but is further explored from the following aspects: 1) Instead of performing input-aware gating, RepMode only uses the task embedding for gating, aiming to adjust its behavior for a specified task;
2) The experts of RepMode can be combined together, which can efficiently utilize multiple experts in an MoE-inspired architecture.

\textbf{Structural re-parameterization.} 
Different from other re-parameterization (re-param) methods \cite{kingma2013auto, zagoruyko2017diracnets, salimans2016weight, figurnov2018implicit}, structural re-param \cite{ding2021repvgg, ding2021diverse} is a recent technique of equivalently converting multi-branch network structures.
With this technique, multi-branch blocks \cite{ding2019acnet, ding2021repvgg, ding2021diverse, wang2022repsr} are introduced to plain networks for enhancing their performance.
However, these methods only achieve inference-time converting, resulting in non-negligible training costs.
There are previous works \cite{ding2022re, hu2022online} accomplishing training-time converting, but they require model-specific optimizer modification \cite{ding2022re} or extra parameters \cite{hu2022online} and only explore its potential on one single task.
In this work, we elegantly incorporate task-specific gating into structural re-param to achieve both training- and inference-time converting for handling multiple tasks, which is more cost-friendly and with better applicability.  
Besides, dynamic convolutions \cite{chen2020dynamic, yang2019condconv, zhang2020dynet,li2021revisiting} also can be roughly considered as re-param methods, which aim to assemble convolutions with the same shape in an input-dependent way.
In contrast, using task-dependent gating, our RepMode can combine experts with diverse configurations to generate composite convolutional kernels, and thus is with higher flexibility to model more situations.


\section{Methodology}

\subsection{Problem definition}  
\label{sec:problem_def}

We start by giving a formal definition of SSP. 
Following \cite{ounkomol2018label}, we assume that each image has \textit{only one} fluorescent label, which greatly relaxes the annotation requirement and makes the setting of this task more general and challenging.
Let $\mathcal{D}=\{(\mathbf{x}_n,\mathbf{y}_n, l_n)\}_{n=1}^{N}$ denotes a SSP dataset with $N$ samples. 
The $n$-th image $\mathbf{x}_n \in \mathcal{I}_n$ is associated with the label $\mathbf{y}_n \in \mathcal{I}_n$,
where $\mathcal{I}_n = \mathbbm{R}^{D_n \times H_n \times W_n}$ denotes the image space and $D_n \times H_n \times W_n$ is the image size. 
The label indicator $l_n \in \mathcal{L} = \{1, 2, ..., S\}$ represents that $\mathbf{y}_n$ is the label of the $l_n$-th subcellular structure, where $S$ is the total number of subcellular structure categories. 
In this work, our goal is to learn a network $F: \mathcal{I} \times \mathcal{L} \to \mathcal{I}$ with the parameters $\boldsymbol\theta$ from $\mathcal{D}$.
SSP can be considered as a collection of $S$ single-label prediction tasks, each of which corresponds to one category of subcellular structures.
To solve SSP, Multi-Net and Multi-Head divide task-specific parameters from $\boldsymbol\theta$ for each task.
In contrast, RepMode aims to share $\boldsymbol\theta$ with all tasks and dynamically organize $\boldsymbol\theta$ to handle specified tasks. 

\subsection{Network architecture}  
\label{sec:net_arc}

The backbone of RepMode is a 3D U-shape encoder-decoder architecture mainly constructed of the downsampling and upsampling blocks.
Specifically, each downsampling block contains two successive MoDE blocks to extract task-specific feature maps and double their channel number, followed by a downsampling layer adopting a convolution with a kernel size of $2 \times 2 \times 2$ and a stride of $2$ to halve their resolution. 
Note that batch normalization (BN) and ReLU activation are performed after each convolutional layer. 
In each upsampling block, an upsampling layer adopts a transposed convolution with a kernel size of $2 \times 2 \times 2$ and a stride of $2$ to upsample feature maps and halve their channel number.
Then, the upsampled feature maps are concatenated with the corresponding feature maps passed from the encoder, and the resultant feature maps are further refined by two successive MoDE blocks.
Finally, a MoDE block without BN and ReLU is employed to reduce the channel number to $1$, aiming to produce the final prediction.
We adopt such a common architecture to highlight the applicability of RepMode and more details are provided in \cref{sec:arc_detail}.
Notably, MoDE blocks are employed in both the encoder and decoder, which can facilitate task-specific feature learning and thus helps to achieve superior performance.

\subsection{Mixture-of-Diverse-Experts block} 
\label{sec:Mode_block}

To handle various prediction tasks of SSP, the representational capacity of the network should be strengthened to guarantee its generalization ability.
Thus, we propose the MoDE block, a powerful alternative to the vanilla convolutional layer, to serve as the basic network component of RepMode.
In the MoDE block, diverse experts are designed to explore a unique convolution collocation, and the gating module is designed to utilize the task-aware prior to produce gating weights for dynamic parameter organizing. 
We delve into the details of these two parts in the following.

\textbf{Diverse expert design.} 
In the MoDE block, we aim to achieve two types of expert diversity:
1) \textit{Shape diversity}: To tackle the multi-scale issue, the experts need to be equipped with various receptive fields;
2) \textit{Kernel diversity}: Instead of irregularly arranging convolutions, it is a better choice to explore a simple and effective pattern to further enrich kernel combinations.
Given these guidelines, we propose to construct \textit{expert pairs} to constitute the multi-branch topology of the MoDE block.
The components of an expert pair are 3D convolutions (Conv) and 3D average poolings (Avgp).
Specifically, an expert pair contains a Conv $K \times K \times K$ expert and an Avgp $K \times K \times K$ - Conv $1 \times 1 \times 1$ expert, 
and we utilize a stride of $1$ and same-padding to maintain the resolution of feature maps.
Overall, the MoDE block is composed of expert pairs with three receptive fields to attain shape diversity (see \cref{fig:pair_and_kernel}(a)).
When $K=1$, since these two experts are equal, only one is preserved for simplicity.
Notably, the Avgp $K \times K \times K$ - Conv $1 \times 1 \times 1$ expert is essentially a special form of the Conv $K \times K \times K$ expert.
To be specific, merging the serial Avgp $K \times K \times K$ kernel and Conv $1 \times 1 \times 1$ kernel would result in a Conv kernel with limited degrees of freedom (named as A-Conv).
Compared to normal Conv, A-Conv has only one learnable parameter and thus acts like a learnable average pooling, which enriches kernel diversity in the same shape (see \cref{fig:pair_and_kernel}(b)). 
The combination of Conv and Avgp is also widely adopted in previous works \cite{ding2021diverse, hu2022online}, but we further explore such a characteristic from the perspective of serial merging.

\input{figures/3-pair_and_kernel}

\textbf{Gating module design.} 
In order to perform a specified single-label prediction task, the task-aware prior needs to be encoded into the network, so that it can be aware of which task is being handled and adjust its behavior to focus on the desired task.
Instead of embedding the task-aware prior by a hash function \cite{dmitriev2019learning} or a complicated learnable module \cite{sun2021task}, we choose the most simple way, \ie embed the task-aware prior of each input image $\mathbf{x}_n$ with the label indicator $l_n$ into a $S$-dimensional one-hot vector $\mathbf{p}_n$, which is expressed as
\begin{equation}
p_{ns} =
\left\{
\begin{aligned}
    1, 
    \quad & \text{if} \ s = l_n, \\
    0, 
    \quad & \text{otherwise},
\end{aligned}
\right.
\quad s = 1, 2, ..., S,
\label{eq:task_emb}
\end{equation}
where $p_{ns}$ indicates the $s$-th entry of $\mathbf{p}_n$.
Then, the task embedding $\mathbf{p}_n$ is fed into the gating module and the gating weights $\mathbf{G}$ are generated by a single-layer fully connected network (FCN) $\phi(\cdot)$, shown as
$
\mathbf{G} = \phi(\mathbf{p}_n) = \{ \mathbf{g}_t \}^T_{t=1}
$ where $T=5$.
Note that we omit $n$ in $\mathbf{G}$ for brevity.
Here $\mathbf{g}_t \in  \mathbbm{R}^{C_\text{O}}$ represents the gating weights for the $t$-th experts, which is split from $\mathbf{G}$, and $C_\text{O}$ is the channel number of the output feature maps.
Finally, $\mathbf{G}$ would be further activated as $\mathbf{\hat{G}} = \{ \mathbf{\hat{g}}_t \}^T_{t=1}$ by Softmax for the balance of the intensity of different experts, which can be formulated as
\begin{align}
  \hat{g}_{ti} = 
  \frac{\exp{(g_{ti})}}{\sum^{T}_{j=1} \exp{(g_{ji})}},
  \quad i = 1, 2, ..., C_\text{O},
  \label{eq:softmax}
\end{align}
where $g_{ti}$ (resp. $\hat{g}_{ti}$) is the $i$-th entry of $\mathbf{g}_t$ (resp. $\mathbf{\hat{g}}_t$).
With the resultant gating weights $\mathbf{\hat{G}}$, RepMode can perform dynamic parameter organizing for these task-agnostic experts conditioned on the task-aware prior.

\input{figures/4-expert_utilization}

\subsection{Gating re-parameterization} 
\label{sec:GatRep}

In addition to studying expert configurations, how to efficiently utilize multiple experts is also worth further exploration.
The traditional manner is to completely utilize all experts to process the input feature maps \cite{ma2018modeling, tang2020progressive, dai2021generalizable, pavlitskaya2020using} (see \cref{fig:expert_utilization}(a)).
However, the output of all experts needs to be calculated and stored,
which would slow down training and inference and increase the GPU memory utilization \cite{ding2021repvgg, zhang2022deep}.
The advanced one is to sparsely route the input feature maps to specific experts \cite{shazeer2017outrageously, hazimeh2021dselect, jacobs1991adaptive, riquelme2021scaling} (see \cref{fig:expert_utilization}(b)).
However, only a few experts are utilized and the others remain unused, which would inevitably reduce the representational capacity of MoE.
To avoid these undesired drawbacks and meanwhile preserve the benefits of MoE, we elegantly introduce task-specific gating to structural re-param, and thus propose GatRep to adaptively fuse the kernels of experts in the MoDE block, through which only one convolution operation is explicitly required (see \cref{fig:expert_utilization}(c)).

\textbf{Preliminary.}
GatRep is implemented based on the homogeneity and additivity of Conv and Avgp, which are recognized in \cite{ding2021diverse, ding2021repvgg}. 
The kernels of a Conv with $C_\text{I}$ input channels, $C_\text{O}$ output channels, and $K \times K \times K$ kernel size is a fifth-order tensor $\mathbf{W} \in \mathcal{Z}(K) = \mathbbm{R}^{C_\text{O} \times C_\text{I} \times K \times K \times K}$, where $\mathcal{Z}(K)$ denotes this kernel space. 
Besides, the kernels of an Avgp $\mathbf{W}^\text{a} \in \mathbbm{R}^{C_\text{I} \times C_\text{I} \times K \times K \times K}$ can be constructed by
\begin{equation}
\mathbf{W}^\text{a}_{c_\text{I}, c^\prime_\text{I}, :, :, :} =
\left\{
\begin{aligned}
    & \frac{1}{K^3}, & 
    \quad \text{if} \ c_\text{I} = c^\prime_\text{I}, \\
    & 0, &
    \quad \text{otherwise},
\end{aligned}
\right.
\label{eq:avgp_kernel}
\end{equation}
where $c_\text{I}, c^\prime_\text{I}, :, :, :$ is the indexes of the tensor and $c_\text{I}, c^\prime_\text{I} = 1, ..., C_\text{I}$.
Note that the Avgp kernel is fixed and thus unlearnable. 
Moreover, we omit the biases here as a common practice. 
Specifically, GatRep can be divided into two steps, \ie serial merging and parallel merging (see \cref{fig:GatRep}).

\textbf{Step 1: serial merging.}
The first step of GatRep is to merge Avgp and Conv into an integrated kernel.
For brevity, here we take an Avgp - Conv expert as an example. 
Let $\mathbf{M}^\text{I}$ denote the input feature maps.
The process of producing the output feature maps $\mathbf{M}^\text{O}$ can be formulated as
\begin{equation}
    \mathbf{M}^\text{O}
    = \mathbf{W} \circledast (\mathbf{W}^\text{a} \circledast \mathbf{M}^\text{I}),
\label{eq:avgp_and_conv}
\end{equation}
where $\circledast$ denotes the convolution operation.
According to the associative law, we can perform an equivalent transformation for \cref{eq:avgp_and_conv} by first combining $\mathbf{W}^\text{a}$ and $\mathbf{W}$. Such a transformation can be expressed as
\begin{equation}
    \mathbf{M}^\text{O}
    = \underbrace{(\mathbf{W} \circledast \mathbf{W}^\text{a})}
        _{\mathbf{W}^\text{e}} 
      \circledast 
      \mathbf{M}^\text{I},
\label{eq:GatRep_step1}
\end{equation}
which means that we first adopt $\mathbf{W}$ to perform a convolution operation on $\mathbf{W}^\text{a}$, and then use the resultant kernel $\mathbf{W}^\text{e}$ to process the input feature maps. 
With this transformation, the kernels of Avgp and Conv can be merged as an integrated one for the subsequent step.

\input{figures/5-GatRep}

\textbf{Step 2: parallel merging.}
The second step of GatRep is to merge all experts in a task-specific manner.
We define $\text{Pad}(\cdot, K^\prime)$ as a mapping function equivalently transferring a kernel to the kernel space $\mathcal{Z}(K^\prime)$ by zero-padding, and set $K^\prime=5$ which is the biggest receptive field size of these experts.
Let $\hat{\mathbf{M}}^\text{O}$ denote the final task-specific feature maps. This transformation can be formulated as
\begin{align}
  \hat{\mathbf{M}}^\text{O} = 
  \underbrace{ \left (
  \sum^T_{t=1} \mathbf{\hat{g}}_t \odot \text{Pad}(\mathbf{W}^\text{e}_t, K^\prime)
  \right )}
  _{\hat{\mathbf{W}}^\text{e}}
  \circledast \mathbf{M}^\text{I},
  \label{eq:GatRep_step2}
\end{align}
where $\odot$ denotes the channel-wise multiplication and $\mathbf{W}^\text{e}_t$ is the kernel of the $t$-th expert. 
Note that $\mathbf{W}^\text{e}_t$ is an integrated kernel (resp. Conv kernel) for an Avgp - Conv expert (resp. Conv expert).
To be specific, the detailed pixel-level form is provided in \cref{sec:pixel_level}.
Finally, $\hat{\mathbf{W}}^\text{e}$ is the resultant task-specific kernel dynamically generated by GatRep.

\input{tables/1-benchmark_results}


\section{Experiments}

\subsection{Experimental setup}
\label{sec:exp_setup}

We conduct the experiments based on the following experimental setup unless otherwise specified.
Due to space limitations, more details are included in \cref{sec:setup_details}.

\textbf{Datasets.} 
For a comprehensive comparison, the dataset is constructed from a dataset collection \cite{ounkomol2018label} containing twelve partially labeled datasets, each of which corresponds to one category of subcellular structures (\ie one single-label prediction task).
All images are resized to make each voxel correspond to $0.29 \times 0.29 \times 0.29$ $\upmu$m$^3$. 
Moreover, we perform per-image z-scored normalization for voxels to eliminate systematic differences in illumination intensity.
For each dataset, we randomly select $25\%$ samples for evaluation and then withhold $10\%$ of the rest for validation.

\textbf{Implementation details.} 
Mean Squared Error (MSE) is adopted as the loss function, which is commonly used to train a regression model. 
Besides, Adam \cite{kingma2014adam} is employed as the optimizer with a learning rate of $0.0001$.
Each model is trained for $1000$ epochs from scratch and validation is performed every $20$ epochs.
Finally, the validated model that attains the lowest MSE is selected for evaluation on the test set.
In a training epoch, we randomly crop a patch with a size of $32 \times 128 \times 128$ from each training image as the input with a batch size of $8$, and random flip is performed for data augmentation.
In the inference stage, we adopt the Gaussian sliding window strategy \cite{isensee2018nnu} to aggregate patch-based output for a full prediction.
To ensure fairness, the same backbone architecture, training configuration, and inference strategy are applied to all comparing models.

\textbf{Evaluation metrics.} 
In addition to MSE, Mean Absolute Error (MAE) and Coefficient of Determination ($R^2$) are also used as the evaluation metrics.
MAE measures absolute differences and thus is less sensitive to outliers than MSE.
$R^2$ measures correlations by calculating the proportion of variance in a label that can be explained by its prediction.
For a clear comparison, we also present the relative overall performance improvement over Multi-Net (\ie $\Delta_{\text{Imp}}$).

\subsection{Comparing to state-of-the-art methods}
\label{sec:comp_sota}

We compared our RepMode to the following methods:
1) Multi-Net: \cite{ounkomol2018label};
2) Multi-Head: include two variants, \ie multiple task-specific decoders (denoted by Dec.) or last layers (denoted by Las.);
3) CondNet \cite{dmitriev2019learning} and TSNs \cite{sun2021task}: two SOTA task-conditional networks for multi-task learning;
4) PIPO-FAN \cite{fang2020multi}, DoDNet \cite{zhang2021dodnet}, and TGNet  \cite{wu2022tgnet}: three SOTA methods of a similar task, \ie partially labeled multi-organ and tumor segmentation (note that DoDNet and TGNet also adopt task-conditioning strategies).

The experimental results on twelve tasks of SSP are reported in \cref{tab:benchmark}.
As recognized in \cite{ounkomol2018label}, the performance of Multi-Net is sufficient to assist with biological research in some cases, thus it can be a reference of reliable metric values for real-life use.
Furthermore, two Multi-Head variants can achieve better performance, which verifies the importance of learning from the complete dataset. 
Notably, PIPO-FAN is an improved Multi-Head variant that additionally constructs a pyramid architecture to handle the multi-scale issue.
The results show that such an architecture can further improve performance but still can not address this issue well.
Moreover, the competitive performance of CondNet and TSNs demonstrates that adopting an appropriate task-conditioning strategy is beneficial. 
However, these networks remain Multi-Head variants since multiple task-specific heads are still required. 
As an advanced version of DoDNet, TGNet additionally modulates the feature maps in the encoder and skip connections, leading to more competitive performance.
It can be observed that, SSP is an extremely tough task since it is hard to attain a huge performance leap in SSP, even for those powerful SOTA methods of related tasks.
However, the proposed RepMode, which aims to learn the task-specific combinations of diverse task-agnostic experts, outperforms the existing methods on ten of twelve tasks of SSP and achieves SOTA overall performance.
Notably, RepMode can even achieve $7.209\%$ (resp. $4.482\%$, $9.176\%$) $\Delta_{\text{Imp}}$ on MSE (resp. MAE, $R^2$), which is near twice the second best method (\ie TGNet).

\input{tables/2-ablation_study}

\subsection{Ablation studies}
\label{sec:ablation}

To verify the effectiveness of the proposed RepMode, we conduct comprehensive ablation studies totally including four aspects, where the results are reported in \cref{tab:ablation_study}.

\textbf{Scope of use of the MoDE block.}
As we mentioned in \cref{sec:net_arc}, we employ the MoDE block in both the encoder and decoder of the network. 
Therefore, we change its scope of use to explore its influence on performance.
The results show that employing it only in the encoder can achieve better performance than only in the decoder, since the encoder can extract task-specific features and pass them to the decoder through skip connections.
Moreover, employing it in both the encoder and decoder is superior since the whole network can perform dynamic parameter organizing.

\input{figures/6-gat_weights}

\textbf{Effectiveness of the expert design.}
The MoDE block is composed of three expert pairs, each of which contains a Conv expert and an Avgp - Conv expert. 
It can be observed that removing experts (especially the Conv experts) from the MoDE block could cause a performance drop due to the degradation of the representational capacity.
Moreover, it is also an interesting finding that the expert pair with the commonly used $3 \times 3 \times 3$ receptive field is most critical.

\input{figures/7-qual_results}

\textbf{Average poolings matter.}
Avgp is one of the basic components of the MoDE block.
The results show that removing such an unlearnable component could also reduce performance, which further verifies the effectiveness of the expert design. 
Besides, we can observe that setting the receptive fields of all Avgp to the same one could also cause a performance drop. This is because being equipped with different receptive fields could facilitate expert diversity.

\textbf{Different gating strategies.}
For task-specific gating, the one-hot task embedding is fed into the single-layer FCN followed by Softmax activation.
Accordingly, we conduct the following modifications:
1) Use the task embedding with each entry sampled from $\mathcal{N}(0, 1)$; 
2) Use the two-layer FCN with the hidden unit number set to $6$;
3) Use Sigmoid activation;
4) Input-dependent gating: input feature maps are first processed by a global average pooling and then fed into the gating module.
The superior performance and simplicity  of the original gating approach demonstrate the applicability of RepMode. 
Notably, input-dependent gating underperforms since an all-shared network can not be aware of the desired task of input without access to any priors.

\subsection{Further analysis}
\label{sec:further_analysis}
In this subsection, we perform further analysis of RepMode to further reveal its capability.
Additional analysis and discussion are provided in \cref{sec:add_anal}.

\textbf{Gating weights visualization.}
In the MoDE block, the gating weights are produced for dynamic parameter organizing, through which the preference of each task for diverse experts can be learned.
As shown in \cref{fig:gat_weights}, the cell membrane relatively prefers the Conv $5 \times 5 \times 5$ expert while the mitochondrion relatively prefers the Conv $1 \times 1 \times 1$ one as we expect.
Besides, the preference of the mid-scale structures (\ie nucleolus and nuclear envelope) is more variable.
Notably, the Avg - Conv experts also be assigned sufficient weights, which verifies the effectiveness of the expert pairs.
It can also be observed that the network pays more attention to small-scale features in the decoder, which could be due to the need for producing a detailed prediction. 

\input{figures/8-TIL_arc}

\textbf{Qualitative results.}
Subcellular structures are hard to be distinguished in transmitted-light images (see \cref{fig:qual_results}).
The second best method (\ie TGNet) suffers from incomplete (see \cref{fig:qual_results}(a)) and redundant (see \cref{fig:qual_results}(b)) predictions, and even yields inexistent patterns (see \cref{fig:qual_results}(c)).
But relatively, RepMode can produce more precise predictions for various subcellular structures at multiple scales even though there are some hard cases (see \cref{fig:qual_results}(a)\&(b)). 
Such a practical advance is crucial in biological research, since inaccurate predictions at some key locations may mislead biologists into making incorrect judgments.

\textbf{RepMode as a better task-incremental learner.}
For a well-trained RepMode, we fine-tuned a newly-introduced expert and gating module for each MoDE block with the other experts frozen, aiming to extend it to an unseen task (see \cref{sec:detailed_TIL} for more details).
As we expect, RepMode can preserve and transfer its domain knowledge through the pretrained experts, which helps to achieve better performance compared to the plain networks (see \cref{tab:TIL}).
As long as the previous gating weights have been stored, such a task-incremental learning manner would not cause any degradation of the performance on the previous tasks.


\input{tables/3-TIL}

\section{Conclusions}

In this paper, we focus on an under-explored and challenging bioimage problem termed SSP, which faces two main challenges, \ie partial labeling and multi-scale.
Instead of constructing a network in a traditional manner, we choose to dynamically organize network parameters with task-aware priors and thus propose RepMode.
Experiments show that RepMode can achieve SOTA performance in SSP.
We believe that RepMode can serve as a stronger baseline for SSP and help to motivate more advances in both the biological and computer vision community.


\section*{Acknowledgements}
We would like to thank Danruo Deng, Bowen Wang, and Jiancheng Huang for their valuable discussion and suggestions.
This work is supported by 
the National Key R\&D Program of China (2022YFE0200700), 
the National Natural Science Foundation of China (Project No. 62006219 and No. 62072452), 
the Natural Science Foundation of Guangdong Province (2022A1515011579), 
the Regional Joint Fund of Guangdong under Grant 2021B1515120011,
and the Hong Kong Innovation and Technology Fund (Project No. ITS/170/20 and ITS/241/21).


{\small
\bibliographystyle{ieee_fullname}
\bibliography{references}
}


\clearpage

\appendix

{\noindent \Large \textbf{Appendix}}

\section{Details of the Network Architecture}
\label{sec:arc_detail}

We have introduced the network architecture of RepMode in \cref{sec:net_arc}.
To guarantee reproducibility, we provide more details in this section.
As shown in \cref{fig:net_arc}, the encoder-decoder architecture of RepMode is mainly constructed of the symmetrical downsampling and upsampling blocks.
Moreover, between the downsampling and upsampling blocks, two successive MoDE blocks are employed to further refine the feature maps.
Finally, a MoDE block without BN and ReLU is used to produce predictions.
It is worth noting that, using the proposed MoDE block and GatRep, any plain network designed for dense prediction tasks can obtain the powerful capability to handle multiple tasks and meanwhile maintain the original architecture, since only the convolutional layers need to be modified.


\section{Pixel-Level Form of GatRep}
\label{sec:pixel_level}

In \cref{sec:GatRep}, we have described the matrix form of GatRep for an intuitive understanding. 
In this section, we provide the pixel-level form as an extension. Note that here we follow the notations described in \cref{sec:GatRep}. 

\textbf{Step1: serial merging.}
In this step, we aim to merge $\mathbf{W}$ and $\mathbf{W}^\text{a}$ of an Avgp - Conv expert into an integrated kernel $\mathbf{W}^\text{e}$. 
This merging is accomplished by using $\mathbf{W}$ to perform a convolution operation on $\mathbf{W}^\text{a}$, formulated as
\begin{align}
    \mathbf{W}^\text{e} & = 
    \mathbf{W} \circledast \mathbf{W}^\text{a},
\end{align}
which is equivalent to
\begin{align}    
    \mathbf{W}^\text{e}_{c_\text{O}, c_\text{I}, d, h, w} & =
    \sum_{i=1}^{C_\text{I}} \mathbf{W}_{c_\text{O}, i, 1, 1, 1} * \mathbf{W}^\text{a}_{i, c_\text{I}, d, h, w},
\end{align}
where the subscripts denote the indexes of tensors in the corresponding dimensions and $*$ is the multiplication.

\textbf{Step 2: parallel merging.}
In this step, we aim to merge the kernels of all experts $\mathbf{W}^\text{e}_t$ where $t = 1, 2, ..., T$.
This merging is accomplished by a linear weighted summation with the gating weights $\mathbf{\hat{G}}_t = \{ \mathbf{\hat{g}}_t \}^T_{t=1}$, formulated as
\begin{align}
    \hat{\mathbf{W}}^\text{e} & = 
    \sum^T_{t=1} \mathbf{\hat{g}}_t \odot \text{Pad}(\mathbf{W}^\text{e}_t, K^\prime),
\end{align}
which is equivalent to
\begin{align}
    \hat{\mathbf{W}}^\text{e}_{c_\text{O}, c_\text{I}, d, h, w} & =
    \sum^T_{t=1} \mathbf{\hat{g}}_{t, c_\text{O}} * 
    \mathbf{W}^\text{p}_{t, c_\text{O}, c_\text{I}, d, h, w},
\end{align}
where $\mathbf{W}^\text{p}$ denotes the kernel processed by $\text{Pad}(\cdot, K^\prime)$.

\input{figures/S9-net_arc}


\section{Details of the Experimental Setup}
\label{sec:setup_details}

In this section, we provide more details of the experimental setup to highlight the comprehensiveness and reproducibility of our experiments.
First, we would provide more descriptions of datasets and implementation details in \cref{sec:supp_ds} and \cref{sec:supp_imp} respectively.
Then, we would provide mathematical definitions of the evaluation metrics in \cref{sec:supp_metrics}.
Finally, we would further describe the comparing state-of-the-art methods in \cref{sec:supp_comp}.

\subsection{Datasets}
\label{sec:supp_ds}

In the experiments, we adopt a dataset collection \cite{ounkomol2018label} to evaluate the performance of the comparing methods and the proposed RepMode in SSP.
The reason why we call it ``dataset collection'' is because it totally contains twelve partially labeled cell image datasets for SSP.
In this dataset collection, each dataset contains $54$ to $80$ high-resolution 3D z-stack image pairs, where each bright-field input is associated with a fluorescent label (as we defined in \cref{sec:problem_def}).
We consolidate these datasets into one single partially labeled dataset to conduct our experiments.
Totally, there are $628$ (resp. $70$, $233$) image pairs for training (resp. validation, test).
With a patch-based training scheme, a dataset of this size is sufficient for such a 3D dense prediction task, which is also recognized by \cite{wu2022tgnet, zhang2021dodnet}.

\subsection{Implementation details}
\label{sec:supp_imp}

All experiments are accomplished with PyTorch 1.12.1 and CUDA 11.6, and run on a single NVIDIA V100 GPU with 32GB memory.
For a fair comparison, all random seeds are fixed at $0$ in each experiment.
Moreover, automatic mixed precision (AMP) is used to accelerate training.
Due to variable image sizes and memory limitations, we adopt a patch-based training scheme in the experiments.
Accordingly, in the validation and test phase, we utilize the Gaussian sliding window strategy \cite{isensee2018nnu} to aggregate patch-based predictions output by the network to obtain the final predictions of full images.
Specifically, we implement the Gaussian sliding window strategy exactly following \cite{zhang2021dodnet} and the window size is set to the same size of training patches (\ie $32 \times 128 \times 128$).

\subsection{Evaluation metrics}
\label{sec:supp_metrics}

The evaluation metrics that we adopted in the experiments include MSE, MAE, and $R^2$.
Following the notations described in \cref{sec:problem_def}, let $\mathbf{y}_n$ and $\mathbf{f}_n$ denote the ground-truth label and the output prediction of $n$-th image pairs respectively.
Furthermore, let $y_{ni}$ and $f_{ni}$ indicate the $i$-th pixel intensity of $\mathbf{y}_n$ and $\mathbf{f}_n$ respectively.
These evaluation metrics can be formulated as
\begin{equation}
    \text{MSE}(\mathbf{y}_n, \mathbf{f}_n) = 
    \frac{1}{P_n}
    \sum_{i=1}^{P_n} (y_{ni} - f_{ni})^2,
\end{equation}
\begin{equation}
    \text{MAE}(\mathbf{y}_n, \mathbf{f}_n) = 
    \frac{1}{P_n}
    \sum_{i=1}^{P_n} \lvert y_{ni} - f_{ni} \rvert,
\end{equation}
\begin{equation}
    R^2(\mathbf{y}_n, \mathbf{f}_n) = 
    1 - 
    \frac{\sum_{i=1}^{P_n} (y_{ni} - f_{ni})^2}{\sum_{i=1}^{P_n} (y_{ni} - \bar{y}_n)^2},
    \label{eq:R2}
\end{equation}
where $P_n$ is the total pixel number of $n$-th image pairs and $\bar{y}_n$ is the average of $y_{ni}$.
We adopt MSE and MAE since they are two commonly used evaluation metrics for regression.
In addition to these two metrics, $R^2$ is also be used in our experiments for two following reasons:
1) Compared to MSE and MAE, $R^2$ further takes into account the variance of the pixel intensity of a ground-truth label (see \cref{eq:R2});
2) MSE and MAE have arbitrary ranges, while $R^2$ normally ranges from $0$ to $1$ and thus is a more intuitive measure.

With these metrics, we report the performance on twelve datasets and present the overall performance by averaging the metrics over all image pairs in \cref{tab:benchmark}.
For a clear comparison, we also report the relative overall performance improvement over Multi-Net which is the most naive baseline.
Let $m_i$ and $m^\prime_i$ denote the overall results of a random method and Multi-Net on the $i$-th metric.
The relative overall performance improvement of this method over Multi-Net on the $i$-th metric can be calculated as
\begin{equation}
    \Delta_{\text{Imp}}(m_i, m^\prime_i) =
    (-1)^{v_i} \frac{m_i - m^\prime_i}{m^\prime_i},
\end{equation}
where $v_i = 1$ if a lower value means better performance for the $i$-th metric, and $0$ otherwise.
With such an informative measure, the performance differences in the experiments can be clearly presented (see \cref{tab:benchmark}). 

\subsection{Comparing methods}
\label{sec:supp_comp}

In \cref{sec:comp_sota}, we have briefly introduced the comparing state-of-the-art methods of the experiments. Here we provide detailed descriptions of these methods:
1) Multi-Net \cite{ounkomol2018label}: multiple individual networks, each of which aims to handle one single-label prediction task; 
2) Multi-Head: a partially-shared network composed of a shared feature extractor and multiple task-specific heads, including two variants, \ie multiple task-specific decoders (denoted by Dec.) or last layers (denoted by Las.);
3) Conditional Network (CondNet) \cite{dmitriev2019learning}: a task-conditional network where the task-aware prior is encoded as feature maps by a predefined hash function;
4) Task Switching Networks (TSNs) \cite{sun2021task}: a task-conditional network that uses a fully connected module to learn the task embedding for adaptive instance normalization;
5) Pyramid Input Pyramid Output Feature Abstraction Network (PIPO-FAN) \cite{fang2020multi}: a network that consists of a U-shape pyramid architecture with multi-resolution images as input, and a deep supervision mechanism to refine the output in different scales;
6) Dynamic On-Demand Network (DoDNet) \cite{zhang2021dodnet}: a task-conditional network composed of a shared encoder-decoder architecture, a controller for filter generation, and a dynamic convolutional head (\ie three convolutional layers);
7) Task-Guided Network (TGNet) \cite{wu2022tgnet}: an improved version of DoDNet, where task-guided residual blocks and attention modules are further introduced to emphasize the features related to the specified task.
Notably, we have equipped these networks with the same backbone of RepMode to ensure fairness.


\section{Additional Analysis and Discussion}
\label{sec:add_anal}

\subsection{Task-incremental learning}
\label{sec:detailed_TIL}

We have conducted the corresponding experiments in \cref{sec:further_analysis} to verify that the proposed RepMode can serve as a better task-incremental learner.
Here we detail the experimental setup and provide additional analysis.

\textbf{Experimental setup.}
We select the mainstream solutions of SSP, \ie Multi-Net and Multi-Head, for a comparison.
For Multi-Head, we select its ``Dec." variant since it contains more task-specific parameters.
First, all these networks are pretrained on eleven datasets. Then, the pretrained networks are extended to a new task by being trained on the remaining dataset.
Note that the training of these two phases also follows the implementation details that we describe in \cref{sec:exp_setup} and \cref{sec:supp_imp}.
Specifically, the strategies of these networks for task-incremental learning are:
1) Multi-Net: employ a new network to be trained on the new dataset from scratch;
2) Multi-Head (Dec.): add a new decoder to handle the new dataset and fine-tunes the whole network;
3) RepMode: introduce an extra expert (here we choose a Conv $3 \times 3 \times 3$ expert) and a new gating module in each MoDE block, and only fine-tune the newly-introduced components with the other ones frozen. 
We adopt the datasets of two basic subcellular structures, \ie nucleolus and cell membrane, for the experiments of task-incremental learning.

\textbf{Results and analysis.}
As shown in \cref{tab:TIL}, the proposed RepMode can achieve superior performance in task-incremental learning.
The main reason is that the experts of RepMode are trained in a task-agnostic manner and thus capable of learning the generalized domain knowledge of SSP.
When trained on a new dataset, RepMode can utilize the pretrained experts to ``transfer'' such knowledge to the new task. 
With this strategy, RepMode can easily adapt to a new task of an unseen subcellular structure, rather than learning it from scratch.
Moreover, as long as the previous gating weights have been stored, the fine-tuned RepMode can maintain the original performance on the previous tasks since the parameters of the frozen experts are fixed and preserved.
Whereas, Multi-Net requires training a new network and thus achieve poor performance in task-incremental learning.
Besides, Multi-Head needs to fine-tune the whole network, which would result in an inevitable performance drop on the previous tasks.

\input{tables/S4-other_blocks}

\subsection{Comparison with other re-param blocks}
\label{sec:supp_rep_blocks}

The performance of the proposed MoDE block is already verified in \cref{sec:ablation}.
In this subsection, we further compare it with the existing SOTA re-param blocks \cite{ding2019acnet, ding2021repvgg, ding2021diverse} in SSP. 
Below we would detail the experimental setup and conduct the corresponding analysis.

\textbf{Experimental setup.}
We select the following state-of-the-art re-param blocks and modify them to a 3D convolution version:
1) Asymmetric convolution network (ACNet) block \cite{ding2019acnet}: consist of a Conv $3 \times 3 \times 3$, a Conv $3 \times 1 \times 3$, and a Conv $3 \times 3 \times 1$;
2) RepVGG block \cite{ding2021repvgg}: contains a Conv $3 \times 3 \times 3$, a Conv $1 \times 1 \times 1$, and a residual connection (since the channel numbers of the input and output feature maps may be different, we replace it with an additional Conv $1 \times 1 \times 1$ aiming to align the channel numbers);
3) Diverse branch block (DBB) \cite{ding2021diverse}: consists of a Conv $1 \times 1 \times 1$, a Conv $1 \times 1 \times 1$ - Conv $K \times K \times K$, a Conv $1 \times 1 \times 1$ - Avgp $K \times K \times K$, and a Conv $K \times K \times K$ (here we set $K=3,5$ and report the best result).
Moreover, in order to adapt to our GatRep for a fair comparison, all BN inside the branches are removed to ensure linearity.
We replace MoDE blocks with these blocks in RepMode, and follow the implementation details that we describe in \cref{sec:exp_setup} and \cref{sec:supp_imp} to evaluate their performance.

\textbf{Results and analysis.}
As we can observe in \cref{tab:other_blocks}, the proposed RepMode can still achieve competitive performance when equipped with different re-param blocks, which reveals its applicability.
Furthermore, compared to the other re-param blocks, our MoDE block can achieve better performance in SSP.
This is because the MoDE block is composed of the experts with diverse configurations.
Such an efficient and flexible convolution collocation works well with a task-conditioning strategy and is capable of handling more generalized situations, which is also demonstrated by the ablation studies in \cref{sec:ablation}.

\input{tables/S5-GatRep_benefit}

\subsection{Cost reducing of GatRep}
In \cref{sec:GatRep}, we have that claimed GatRep is an efficient expert utilization manner for the MoDE block.
Specifically, compared to completely utilizing all
experts to process the input feature maps (see \cref{fig:expert_utilization}(a)), GatRep can significantly reduce the computational and memory costs caused by the multi-branch topology of MoE.
In this subsection, we provide some empirical evidence to demonstrate this benefit of GatRep.
As we can observe in \cref{tab:GatRep_benefit}, GatRep can save $41.02\%$ and $61.26\%$ time in a training epoch and a validation respectively.
This is because only one convolution operation is required in a MoDE block when using GatRep.
Moreover, GatRep can reduce $37.85\%$ peak GPU memory utilization, since the output feature maps of all experts are no longer separately calculated and stored.
Using GatRep, our RepMode can acquire cost-economic performance improvement and the ability to handle multiple tasks in an all-shared network.
As a result, RepMode can maintain a compact practical topology exactly like a plain network, and meanwhile achieves a powerful theoretical topology.
Such a technique can increase the device-friendliness of RepMode in the practical scenarios of biological research.

\input{figures/S10-extra_qual_results}

\subsection{Experimental results of multiple runs}
To further verify the effectiveness of the proposed RepMode, we perform ``four-fold cross-test'' and report the average results of multiple runs.
Specifically, following the ratio of $25\%$, we divide the dataset into four parts and then select each part in turn as the test set to conduct the experiments. 
We compare our RepMode with a Multi-Head variant (\ie Multi-Head (Dec.)) and two competitive methods (\ie TSNs \cite{sun2021task} and TGNet\cite{wu2022tgnet}).
The experimental results show that our RepMode remains superior (see \cref{tab:multiple_runs}).


\input{tables/S6-multiple_runs}

\section{More Qualitative Examples}
\label{sec:more_examples}

In this section, we provide more qualitative examples as an extension to \cref{fig:qual_results}.
It is worth noting that all images, including transmitted-light images, fluorescent images, and prediction results, are visualized by Imaris 9.0.1 with identical rendering configurations respectively for a fair comparison.
Moreover, all examples are randomly selected from the test set, and a random z-axis slice is presented for each example.
As shown in \cref{fig:extra_qual_results}, Our RepMode can produce relatively precise predictions even for those hard cases (\eg \cref{fig:extra_qual_results}(c)\&(d)), which demonstrates the remarkable effectiveness of RepMode in SSP.

\end{document}

%% file: figures/1-SSP.tex
\begin{figure}[t]
    \centering
    \includegraphics[width=1\linewidth]{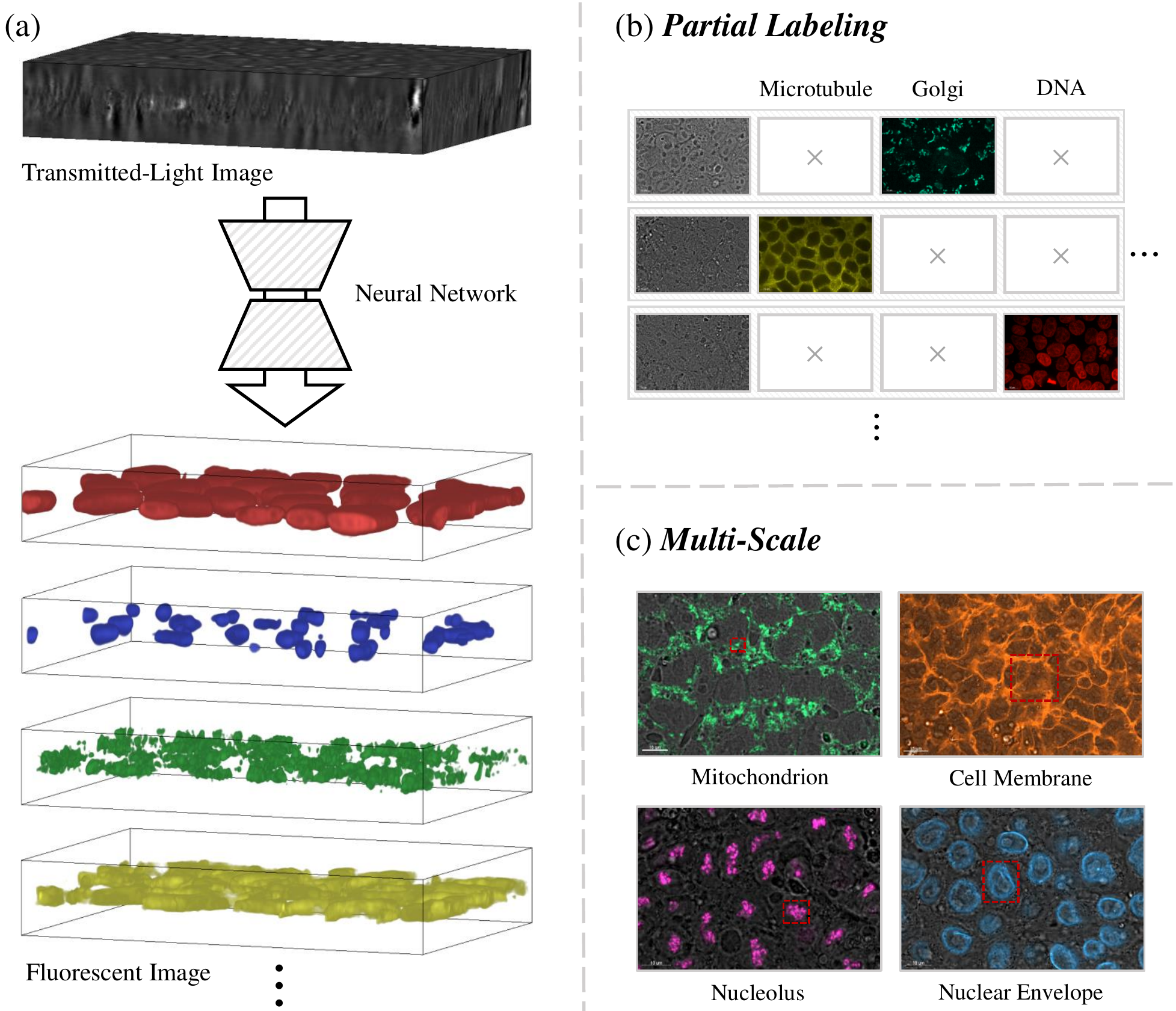}
    \caption{
        (a) Illustration of subcellular structure prediction (SSP), which aims to predict the 3D fluorescent images of multiple subcellular structures from a 3D transmitted-light image. This task faces two challenges, \ie (b) partial labeling and (c) multi-scale.
    }
    \label{fig:SSP}
     \vspace{-3mm}
\end{figure}

%% file: figures/2-overview.tex
\begin{figure*}[t]
    \centering
    \includegraphics[width=1\linewidth]{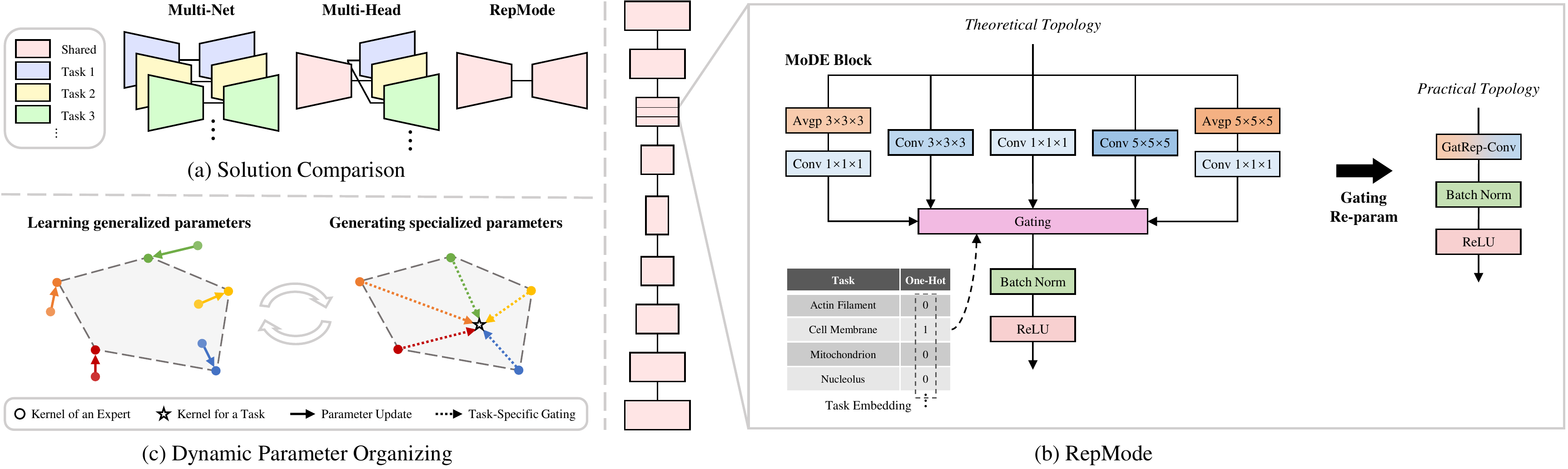}
    \caption{
        Overview of the proposed method. 
        (a) Comparison of two mainstream solutions (\ie Multi-Net and Multi-Head) and the proposed method (\ie RepMode) for SSP.
        (b) Illustration of our RepMode which includes two key components, \ie the proposed MoDE block and GatRep.
        (c) Diagram of how RepMode dynamically organizes its parameters in a MoDE block.
        Note that the gray region denotes the convex hull decided by the expert kernels, and the convex hull is the area where the task-specific kernels would be situated.
    }
    \label{fig:overview}
     \vspace{-3mm}
\end{figure*}

%% file: figures/3-pair_and_kernel.tex
\begin{figure}[t]
    \centering
    \includegraphics[width=1\linewidth]{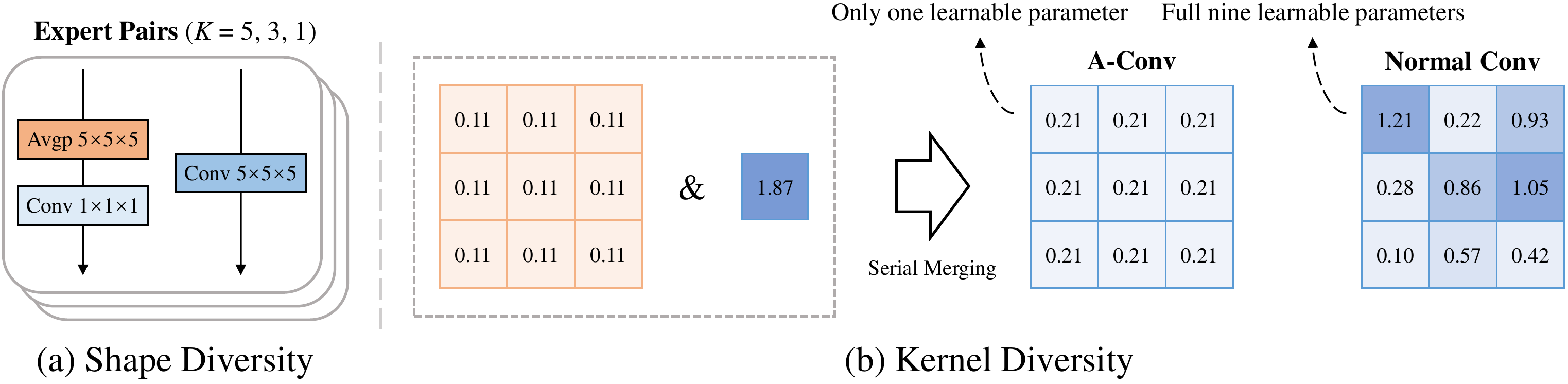}
    \caption{
        (a) Expert pairs with a receptive field size of $K=5,3,1$. Note that each branch denotes an expert.
        (b) Examples of two types of Conv kernels in an expert pair, including A-Conv and normal Conv. Here we present the kernels of a 2D version with a receptive field size of $3$ for simplicity.
    }
    \label{fig:pair_and_kernel}
     \vspace{-3mm}
\end{figure}

%% file: figures/4-expert_utilization.tex
\begin{figure}[t]
    \centering
    \includegraphics[width=1\linewidth]{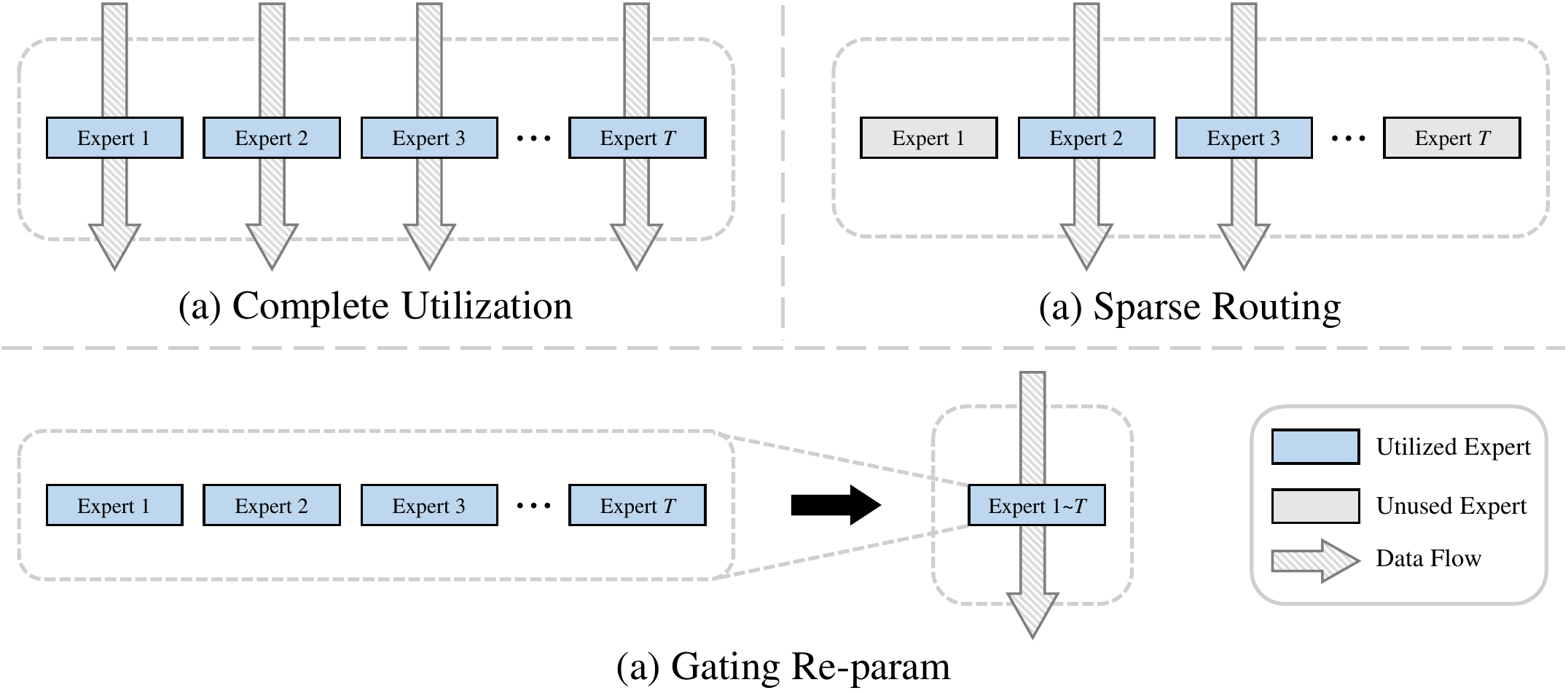}
    \caption{
        Diagram of expert utilization manners of MoE, including
        (a) complete utilization, (b) sparse routing, and (c) our GatRep.
    }
    \label{fig:expert_utilization}
     \vspace{-3mm}
\end{figure}

%% file: figures/5-GatRep.tex
\begin{figure}[t]
    \centering
    \includegraphics[width=1\linewidth]{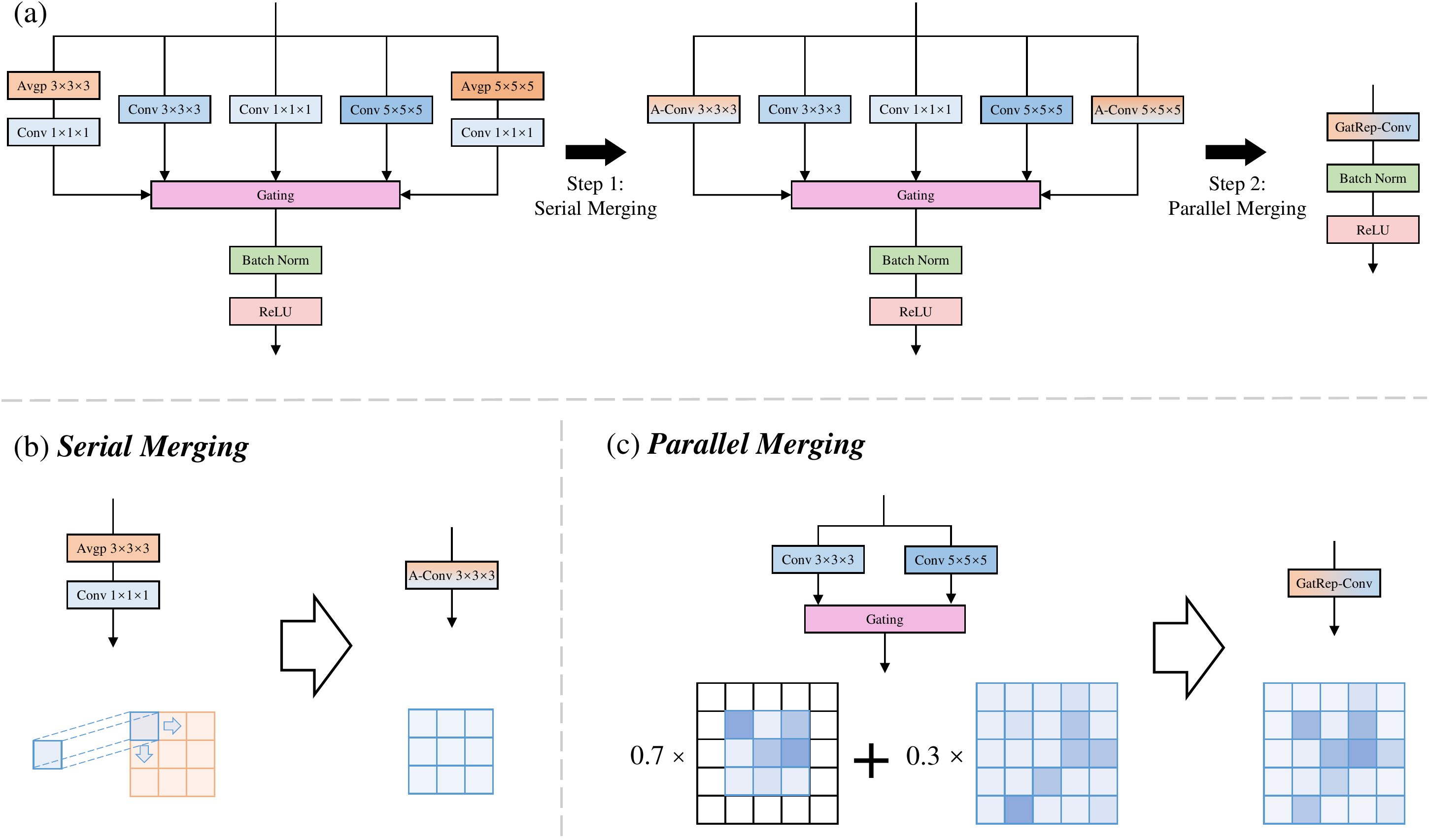}
    \caption{
        (a) Process of GatRep which includes two steps, \ie (b) serial merging and (c) parallel merging.
    }
    \label{fig:GatRep}
     \vspace{-3mm}
\end{figure}

%% file: tables/1-benchmark_results.tex
\begin{table*}[t]
  \centering
  \setlength{\tabcolsep}{1pt}
  \renewcommand{\arraystretch}{1.2}
  \resizebox{\linewidth}{!}{
    \setlength{\tabcolsep}{1mm}{
    \begin{tabular}{l|ccc|ccc|ccc|ccc|ccc|ccc|ccc}
    \hline
    \multirow{2}[3]{*}{Methods} & \multicolumn{3}{c|}{Actin Filament} & \multicolumn{3}{c|}{Actom. Bundle} & \multicolumn{3}{c|}{Cell Membrane} & \multicolumn{3}{c|}{Desmosome} & \multicolumn{3}{c|}{DNA} & \multicolumn{3}{c|}{Endop. Reticulum} & \multicolumn{3}{c}{Golgi Apparatus} \bigstrut[b]\\
\cline{2-22}          & MSE   & MAE   & $R^2$ & MSE   & MAE   & $R^2$ & MSE   & MAE   & $R^2$ & MSE   & MAE   & $R^2$ & MSE   & MAE   & $R^2$ & MSE   & MAE   & $R^2$ & MSE   & MAE   & $R^2$ \bigstrut\\
    \hline
    Multi-Net \cite{ounkomol2018label} & .4241 & .4716 & .5695 & .7247 & .4443 & .2606 & .5940 & .4351 & .3930 & .8393 & .5640 & .0162 & .5806 & .5033 & .3822 & .4635 & .4914 & .5262 & .8023 & .5732 & .0801 \bigstrut[t]\\
    Multi-Head (Dec.) & .4278 & .4803 & .5657 & .7052 & .4363 & .2804 & .5785 & .4625 & .4089 & .8431 & .5677 & .0118 & .5312 & .4764 & .4346 & .4454 & .4832 & .5448 & .7925 & .5768 & .0910 \\
    Multi-Head (Las.) & .4648 & .4978 & .5281 & .6697 & .4222 & .3168 & .5568 & .4441 & .4310 & .8402 & .5637 & .0148 & .5088 & .4824 & .4581 & .4372 & .4697 & .5531 & .7918 & .5807 & .0921 \\
    CondNet \cite{dmitriev2019learning} & .4246 & .4719 & .5688 & .6873 & .4286 & .2988 & .5635 & .4157 & .4242 & .8422 & .5655 & .0126 & .4967 & .4707 & .4712 & .4290 & .4697 & .5615 & .7996 & .5823 & .0831 \\
    TSNs \cite{sun2021task} & .4279 & .4779 & .5656 & .6691 & .4111 & .3174 & \textbf{.5309} & .4346 & \textbf{.4575} & .8392 & .5630 & .0160 & .4974 & .4682 & .4702 & .4362 & .4785 & .5543 & .7892 & .5777 & .0949 \\
    PIPO-FAN \cite{fang2020multi} & .4063 & .4603 & .5873 & .6815 & .4306 & .3046 & .5440 & .4389 & .4441 & .8417 & .5674 & .0131 & .4868 & .4626 & .4813 & .4433 & .4832 & .5470 & .7968 & .5861 & .0861 \\
    DoDNet \cite{zhang2021dodnet} & .4215 & .4706 & .5721 & .6989 & .4204 & .2870 & .5459 & .4390 & .4422 & .8415 & .5633 & .0133 & .5280 & .4810 & .4382 & .4414 & .4844 & .5490 & .7927 & .5774 & .0909 \\
    TGNet \cite{wu2022tgnet} & \textbf{.3917} & \textbf{.4535} & \textbf{.6023} & .6843 & .4213 & .3018 & .5856 & .4227 & .4015 & .8392 & .5654 & .0160 & .5011 & .4746 & .4666 & .4441 & .4806 & .5460 & .7870 & .5774 & .0973 \\
    RepMode & .3936 & .4558 & .6004 & \textbf{.6572} & \textbf{.4103} & \textbf{.3295} & .5443 & \textbf{.4136} & .4437 & \textbf{.8358} & \textbf{.5619} & \textbf{.0199} & \textbf{.4852} & \textbf{.4598} & \textbf{.4831} & \textbf{.4046} & \textbf{.4445} & \textbf{.5865} & \textbf{.7792} & \textbf{.5694} & \textbf{.1064} \bigstrut[b]\\
    \hline
    \end{tabular}%
    }}
  \resizebox{\linewidth}{!}{
    \setlength{\tabcolsep}{1mm}{

    \begin{tabular}{l|ccc|ccc|ccc|ccc|ccc|ccc|ccc}
    \hline
    \multirow{2}[4]{*}{Methods} & \multicolumn{3}{c|}{Microtubule} & \multicolumn{3}{c|}{Mitochondria} & \multicolumn{3}{c|}{Nuclear Envelope} & \multicolumn{3}{c|}{Nucleolus} & \multicolumn{3}{c|}{Tight Junction} & \multicolumn{3}{c|}{All} & \multicolumn{3}{c}{$\Delta_{\text{Imp}} \ (\%)$} \bigstrut\\
\cline{2-22}          & MSE   & MAE   & $R^2$ & MSE   & MAE   & $R^2$ & MSE   & MAE   & $R^2$ & MSE   & MAE   & $R^2$ & MSE   & MAE   & $R^2$ & MSE   & MAE   & $R^2$ & MSE   & MAE   & $R^2$ \bigstrut\\
    \hline
    Multi-Net \cite{ounkomol2018label} & .3682 & .4348 & .6296 & .4684 & .3921 & .5172 & .3014 & .3006 & .6954 & .2164 & .1789 & .7826 & .6474 & .3369 & .3370 & .5341 & .4269 & .4337 & 0.000 & 0.000 & 0.000 \bigstrut[t]\\
    Multi-Head (Dec.) & .3932 & .4594 & .6044 & .4545 & .3888 & .5315 & .2687 & .2895 & .7284 & .2114 & .1762 & .7877 & .6396 & .3252 & .3451 & .5226 & .4258 & .4456 & 2.149 & 0.272 & 2.752 \\
    Multi-Head (Las.) & .3781 & .4465 & .6196 & .4649 & .3991 & .5208 & .2909 & .3057 & .7059 & .2213 & .1870 & .7778 & .6547 & .3367 & .3298 & .5223 & .4275 & .4461 & 2.218 & -0.12 & 2.868 \\
    CondNet \cite{dmitriev2019learning} & .3868 & .4523 & .6108 & .4673 & .4067 & .5184 & .2876 & .3014 & .7094 & .2203 & .1865 & .7787 & .6569 & .3241 & .3274 & .5206 & .4232 & .4478 & 2.534 & 0.884 & 3.249 \\
    TSNs \cite{sun2021task} & .3407 & .4235 & .6572 & .4625 & .3956 & .5233 & .2904 & .2991 & .7064 & .2116 & .1751 & .7874 & .6479 & .3320 & .3367 & .5113 & .4192 & .4572 & 4.263 & 1.804 & 5.437 \\
    PIPO-FAN \cite{fang2020multi} & .3604 & .4365 & .6373 & .4750 & .4171 & .5105 & .2904 & .3003 & .7065 & .2097 & .1782 & .7894 & .6437 & .3282 & .3410 & .5141 & .4237 & .4543 & 3.747 & 0.766 & 4.764 \\
    DoDNet \cite{zhang2021dodnet} & .3972 & .4606 & .6004 & .4772 & .4119 & .5081 & .2976 & .3164 & .6992 & .2250 & .1934 & .7740 & .6703 & .3336 & .3137 & .5276 & .4291 & .4406 & 1.227 & -0.49 & 1.607 \\
    TGNet \cite{wu2022tgnet} & .3569 & .4310 & .6410 & .4585 & .3971 & .5274 & .2748 & .2940 & .7222 & .2093 & .1799 & .7897 & .6232 & \textbf{.3238} & .3619 & .5108 & .4183 & .4578 & 4.363 & 2.022 & 5.566 \\
    RepMode & \textbf{.3389} & \textbf{.4171} & \textbf{.6590} & \textbf{.4459} & \textbf{.3885} & \textbf{.5404} & \textbf{.2631} & \textbf{.2820} & \textbf{.7340} & \textbf{.1995} & \textbf{.1682} & \textbf{.7997} & \textbf{.6168} & .3245 & \textbf{.3685} & \textbf{.4956} & \textbf{.4078} & \textbf{.4735} & \textbf{7.209} & \textbf{4.482} & \textbf{9.176} \bigstrut[b]\\
    \hline
    \end{tabular}%
    }}
    
     \caption{
     Experimental results of the proposed RepMode and the comparing methods on twelve prediction tasks of SSP. 
     The best performance (lowest MSE and MAE, highest $R^2$) is marked in bold.
     Note that ``All'' indicates the overall performance.
     }
      
    \label{tab:benchmark}
    \vspace{-3mm}
\end{table*}

%% file: tables/2-ablation_study.tex
\begin{table}[t]
  \centering
  \setlength{\tabcolsep}{1pt}
  \renewcommand{\arraystretch}{1.2}

  \resizebox{0.97\linewidth}{!}{
    \setlength{\tabcolsep}{2mm}{
    \begin{tabular}{l|l|ccc}
    \hline
    Ablation & Methods & MSE   & MAE   & $R^2$ \bigstrut\\
    \hline
    \multirow{2}[2]{*}{Scope} & only in Dec. & .5097 & .4139 & .4590 \bigstrut[t]\\
          & only in Enc. & .5079 & .4184 & .4607 \bigstrut[b]\\
    \hline
    \multirow{5}[2]{*}{Expert} & w/o $1 \times 1 \times 1$ expert pair & .5027 & .4106 & .4662 \bigstrut[t]\\
          & w/o $3 \times 3 \times 3$ expert pair & .5080 & .4141 & .4605 \\
          & w/o $5 \times 5 \times 5$ expert pair & .5017 & .4108 & .4672 \\
          & w/o Conv expert & .5631 & .4346 & .4042 \\
          & w/o Avgp - Conv expert & .5037 & .4101 & .4651 \bigstrut[b]\\
    \hline
    \multirow{3}[1]{*}{Average Pooling} & w/o Avgp & .4999 & .4112 & .4691 \bigstrut[t]\\
          & all use Avgp $3 \times 3 \times 3$ & .4974 & .4072 & .4716 \\
          & all use Avgp $5 \times 5 \times 5$ & .4964 & .4091 & .4725 \\
    \hline
    \multirow{4}[1]{*}{Gating} & use Gauss. task embedding & .5071 & .4155 & .4616 \\
          & use two-layer FCN & .4980 & .4060 & .4710 \\
          & use Sigmoid activation & .4992 & .4094 & .4698 \\
          & Input-dep. gating & .7958 & .5527 & .1619 \bigstrut[b]\\
    \hline
    Original & RepMode & \textbf{.4956} & \textbf{.4078} & \textbf{.4735} \bigstrut\\
    \hline
    \end{tabular}
    }}
    
    \caption{
        Ablation studies from four aspects.
        Note that ``Methods" denotes the variants of the proposed RepMode.
    }
      
    \label{tab:ablation_study}
    \vspace{-3mm}
\end{table}

%% file: figures/6-gat_weights.tex
\begin{figure}[t]
    \centering
    \includegraphics[width=1\linewidth]{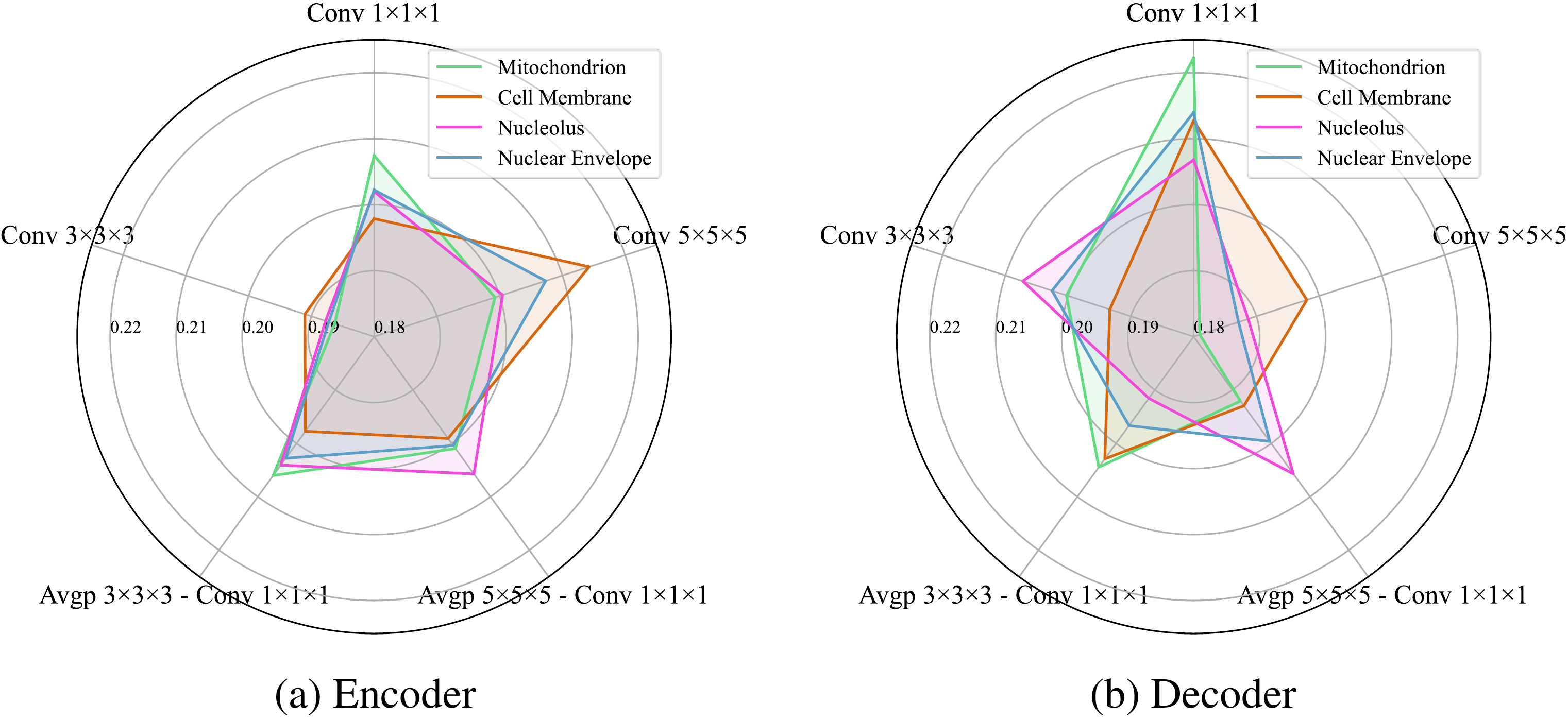}
    \caption{
        Visualization of channel-wisely averaged gating weights of two MoDE blocks, which are randomly selected from the encoder and decoder of a well-trained RepMode, respectively.
        The results of subcellular structures shown in \cref{fig:SSP}(c) are presented.
    }
    \label{fig:gat_weights}
     \vspace{-3mm}
\end{figure}

%% file: figures/7-qual_results.tex
\begin{figure*}[t]
    \centering
    \includegraphics[width=1\linewidth]{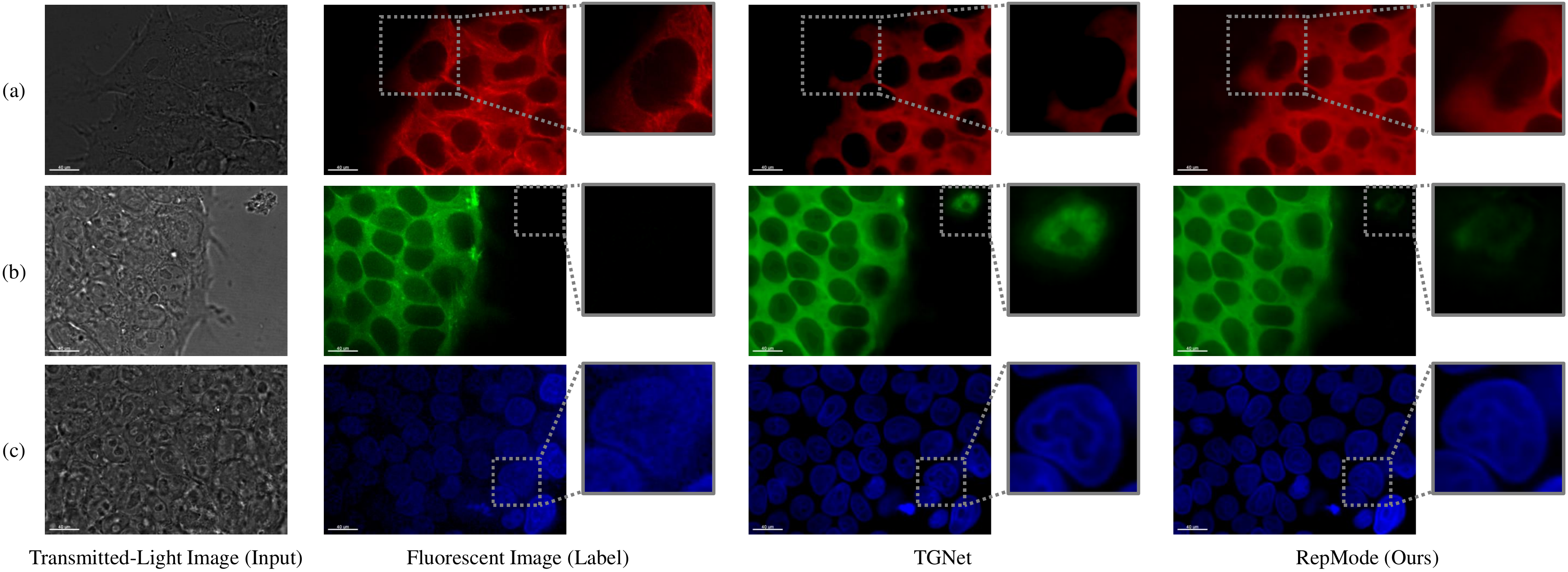}
    \caption{
        Examples of the prediction results on the test set, including
        (a) microtubule, (b) actin filament, and (c) DNA.
        We compare the predictions of our RepMode with the ones of TGNet \cite{wu2022tgnet} which is a competitive method.
        Note that the dotted boxes indicate the major prediction difference. 
        More examples are provided in \cref{sec:more_examples}.
    }
    \label{fig:qual_results}
     \vspace{-3mm}
\end{figure*}

%% file: figures/8-TIL_arc.tex
\begin{figure}[t]
    \centering
    \includegraphics[width=0.92\linewidth]{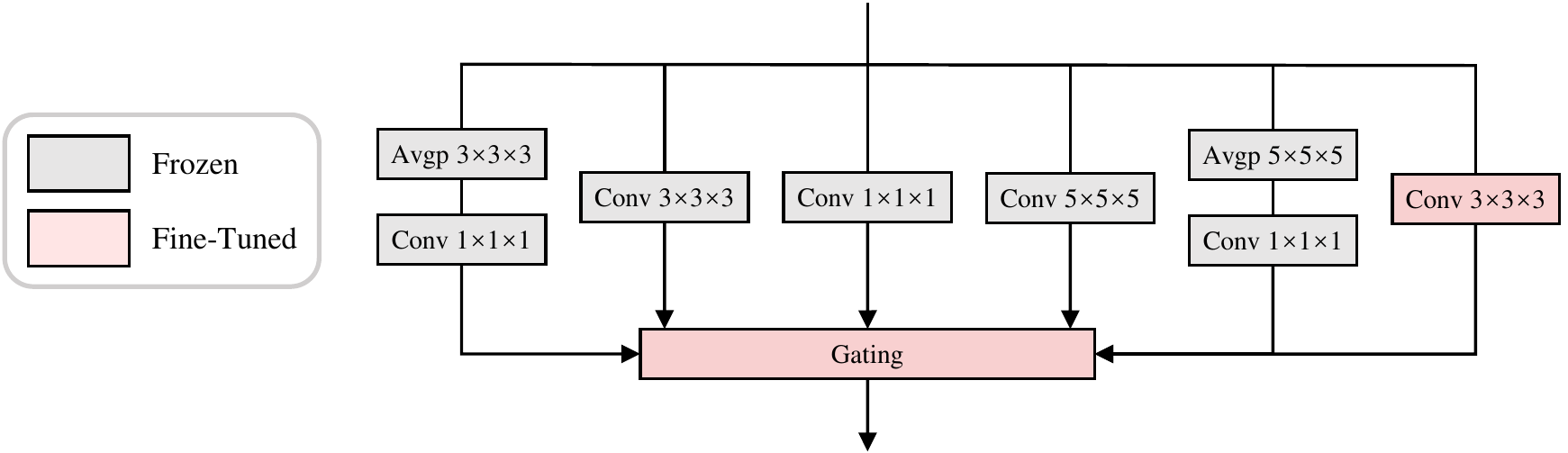}
    \caption{
        Diagram of the MoDE block for task-incremental learning. Note that here we employ an extra Conv $3 \times 3 \times 3$ expert.
    }
    \label{fig:TIL_arc}
     \vspace{-3mm}
\end{figure}

%% file: tables/3-TIL.tex
\begin{table}[t]
  \centering
  \setlength{\tabcolsep}{1pt}
  \renewcommand{\arraystretch}{1.2}

  \resizebox{\linewidth}{!}{
    \setlength{\tabcolsep}{1mm}{
    \begin{tabular}{l|l|ccc|ccc}
    \hline
    \multirow{2}[4]{*}{Methods} & \multirow{2}[4]{*}{Strategies} & \multicolumn{3}{c|}{Nucleolus} & \multicolumn{3}{c}{Cell Membrane} \bigstrut\\
\cline{3-8}          &       & MSE   & MAE   & $R^2$ & MSE   & MAE   & $R^2$ \bigstrut\\
    \hline
    Mutli-Net \cite{ounkomol2018label} & Individual Training & .2164 & .1789 & .7826 & .5940 & .4351 & .3930 \bigstrut[t]\\
    Multi-Head (Dec.) & All Fine-Tuning & .2121 & .1811 & .7870 & .5339 & .4097 & .4543 \\
    RepMode & Experts Frozen & \textbf{.2052} & \textbf{.1774} & \textbf{.7939} & \textbf{.5260} & \textbf{.4077} & \textbf{.4625} \bigstrut[b]\\
    \hline
    \end{tabular}%
    }}
    
    \caption{Experimental results of task-incremental learning for two basic subcellular structures.
    Note that the strategies of the experimental methods for extending to a new task are presented. 
    }
      
    \label{tab:TIL}
     \vspace{-3mm}
\end{table}

%% file: figures/S9-net_arc.tex
\begin{figure}[t]
    \centering
    \includegraphics[width=1\linewidth]{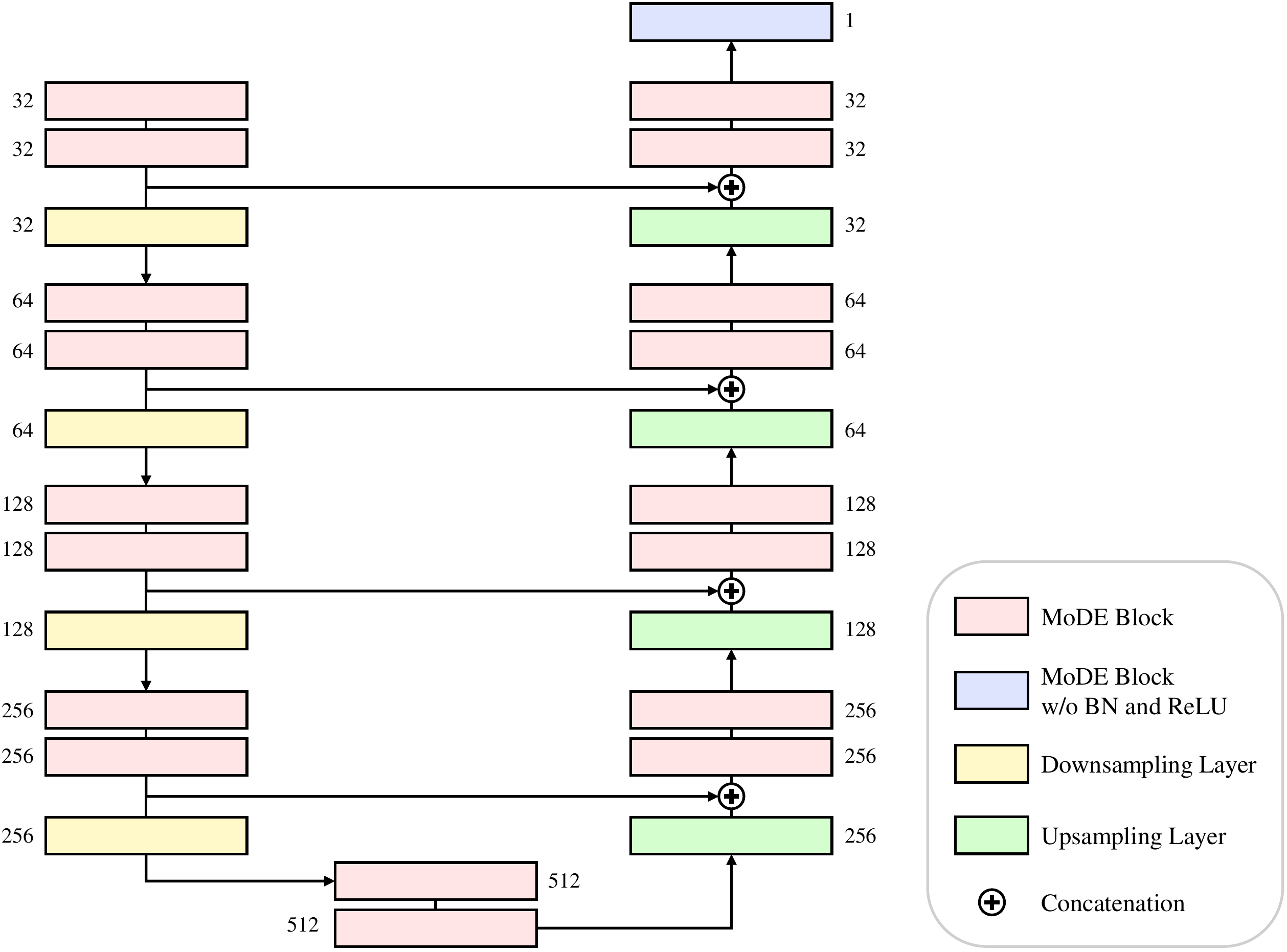}
    \caption{
        Detailed architecture of the proposed RepMode.
        The channel number of the output feature maps is shown next to each block.
        Note that we omit some components (\eg skip connections and the final MoDE block) in \cref{fig:overview} for the sake of brevity.
    }
    \label{fig:net_arc}
     \vspace{-3mm}
\end{figure}

%% file: tables/S4-other_blocks.tex
\begin{table}[t]
  \centering
  \setlength{\tabcolsep}{1pt}
  \renewcommand{\arraystretch}{1.2}

  \resizebox{0.75\linewidth}{!}{
    \setlength{\tabcolsep}{3mm}{
    \begin{tabular}{l|ccc}
    \hline
    Blocks & MSE   & MAE   & $R^2$ \bigstrut\\
    \hline
    ACNet Block \cite{ding2019acnet} & .5075 & .4197 & .4611 \bigstrut[t]\\
    RepVGG Block \cite{ding2021repvgg} & .5034 & .4122 & .4654 \\
    DBB \cite{ding2021diverse} & .5023 & .4102 & .4667 \\
    MoDE Block & \textbf{.4956} & \textbf{.4078} & \textbf{.4735} \bigstrut[b]\\
    \hline
    \end{tabular}%
    }}
    
    \caption{
    Comparison with other SOTA re-param blocks in SSP. 
    Note that we modify these blocks to adapt to our GatRep.
    }
      
    \label{tab:other_blocks}
     \vspace{-3mm}
\end{table}

%% file: tables/S5-GatRep_benefit.tex
\begin{table}[t]
  \centering
  \setlength{\tabcolsep}{1pt}
  \renewcommand{\arraystretch}{1.2}

  \resizebox{1\linewidth}{!}{
    \setlength{\tabcolsep}{2mm}{
    \begin{tabular}{l|cc|c}
    \hline
    \multirow{2}[4]{*}{Methods} & \multicolumn{2}{c|}{Time (s)} & \multicolumn{1}{c}{\multirow{2}[4]{*}{GPU Memory (\%)}} \bigstrut\\
\cline{2-3}          & Training & Validation &  \bigstrut\\
    \hline
    RepMode w/o GatRep & 135.87 & 1359.41 & 95.04 \bigstrut[t]\\
    RepMode w/ GatRep & \textbf{80.13} & \textbf{526.59} & \textbf{59.07} \bigstrut[b]\\
    \hline
    \end{tabular}%
    }}
    
    \caption{
        Statistics of time and memory consumption.
        Note that ``Time'' indicates the average time of a training epoch or a validation in a complete training phase, and ``GPU Memory'' indicates the maximum percentage of allocated GPU memory during training.
        These results are acquired based on an NVIDIA V100 GPU with 32GB memory.
    }
      
    \label{tab:GatRep_benefit}
     \vspace{-3mm}
\end{table}

%% file: figures/S10-extra_qual_results.tex
\begin{figure*}[t]
    \centering
    \includegraphics[width=1\linewidth]{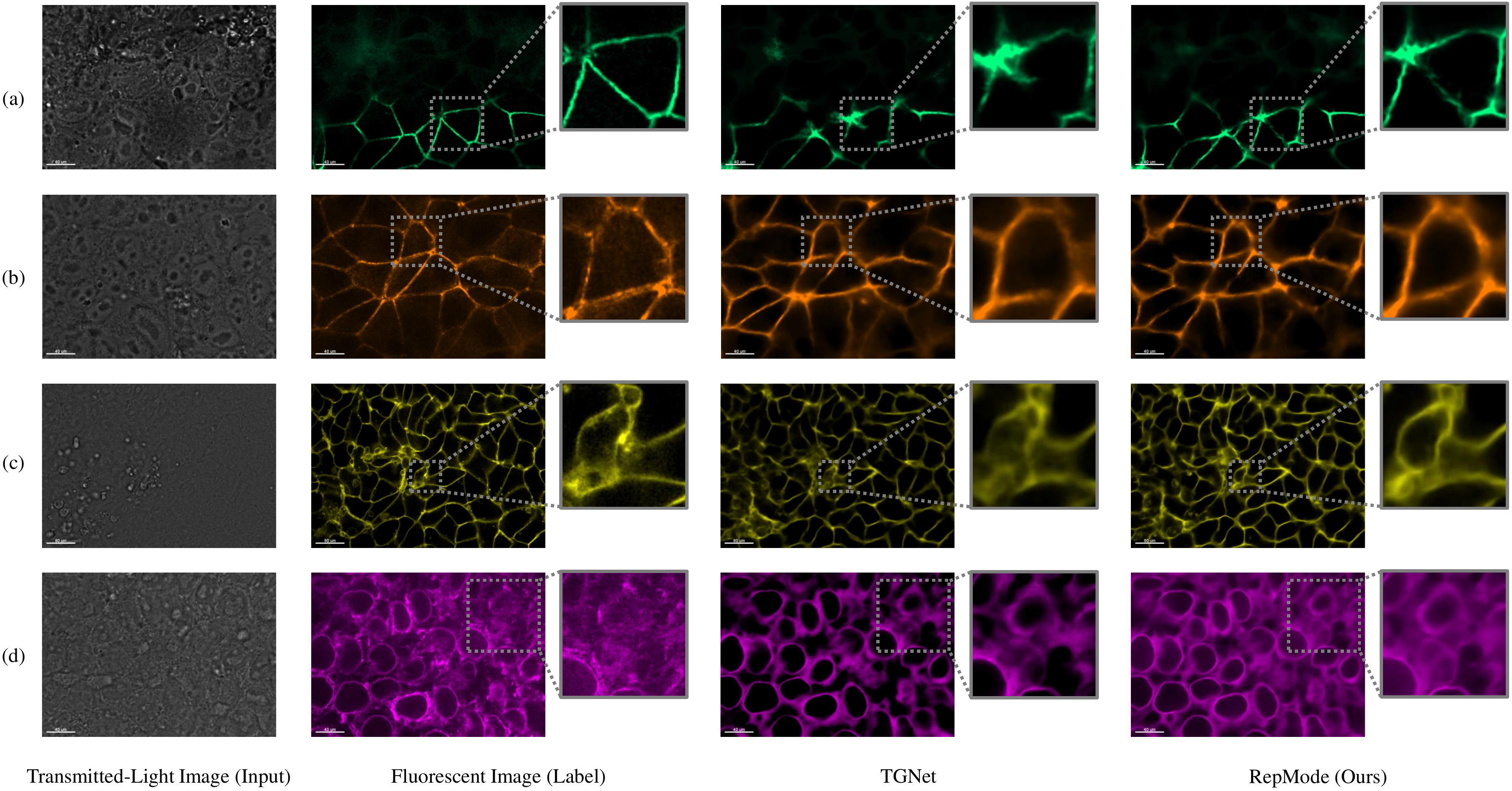}
    \caption{
        Examples of the prediction results on the test set, including
        (a) tight junction, (b) actomyosin bundle, (c) cell membrane, and (d) endoplasmic reticulum.
        We compare the predictions of our RepMode with the ones of TGNet \cite{wu2022tgnet} which is a competitive method.
        Note that the dotted boxes indicate the major prediction difference. 
    }
    \label{fig:extra_qual_results}
\end{figure*}

%% file: tables/S6-multiple_runs.tex
\begin{table}[t]
  \centering
  \setlength{\tabcolsep}{1pt}
  \renewcommand{\arraystretch}{1.2}

  \resizebox{0.77\linewidth}{!}{
    \setlength{\tabcolsep}{3mm}{
    \begin{tabular}{l|ccc}
    \hline
    Methods & MSE   & MAE   & $R^2$ \bigstrut\\
    \hline
    Multi-Head (Dec.) & .5204 & .4247 & .4466 \bigstrut[t]\\
    TSNs \cite{sun2021task} & .5134 & .4202 & .4538 \\
    TGNet \cite{wu2022tgnet} & .5123 & .4186 & .4549 \\
    RepMode & \textbf{.5032} & \textbf{.4124} & \textbf{.4642} \bigstrut[b]\\
    \hline
    \end{tabular}
    }}
    
    \caption{
    Experimental results of ``four-fold cross-test''. Note that we present the average results of multiple runs. 
    }
      
    \label{tab:multiple_runs}
     \vspace{-3mm}
\end{table}